\documentclass[10pt,twocolumn,letterpaper]{article}

\usepackage[pagenumbers]{cvpr} %

\usepackage{comment}
\usepackage{booktabs} %
\usepackage{comment}
\usepackage{tabularx}
\usepackage{multirow}
\usepackage{caption}
\usepackage{subcaption}
\usepackage{dsfont}
\usepackage{mathrsfs}
\usepackage{ftnright}

\usepackage{pifont}%
\newcommand{\xmark}{\ding{55}}%

\usepackage{xspace}
\newcommand{\OURS}{Pix2NPHM\xspace}

\DeclareMathOperator*{\argmin}{arg\,min}

\definecolor{cvprblue}{rgb}{0.21,0.49,0.74}
\usepackage[pagebackref,breaklinks,colorlinks,allcolors=cvprblue]{hyperref}

\def\paperID{XXXXX} %
\def\confName{CVPR}
\def\confYear{2026}

\title{Pix2NPHM: Learning to Regress NPHM Reconstructions From a Single Image}

\author{
Simon Giebenhain$^1$ \quad
Tobias Kirschstein$^1$ \quad
Liam Schoneveld$^2$ \\
 Davide Davoli$^{3*}$ \quad
  Zhe Chen$^2$ \quad
Matthias Nie{\ss}ner$^1$ \vspace{0.2cm}\\
$^1$Technical University of Munich  \qquad $^2$Woven by Toyota \qquad $^3$Toyota Motor Europe
}

\begin{document}

\twocolumn[{
\renewcommand\twocolumn[1][]{#1}
\maketitle
\thispagestyle{empty}
\begin{center}
  \newcommand{\teaserwidth}{\textwidth}
   \vspace{-0.8cm}

  \centerline{

    \includegraphics[width=\teaserwidth]{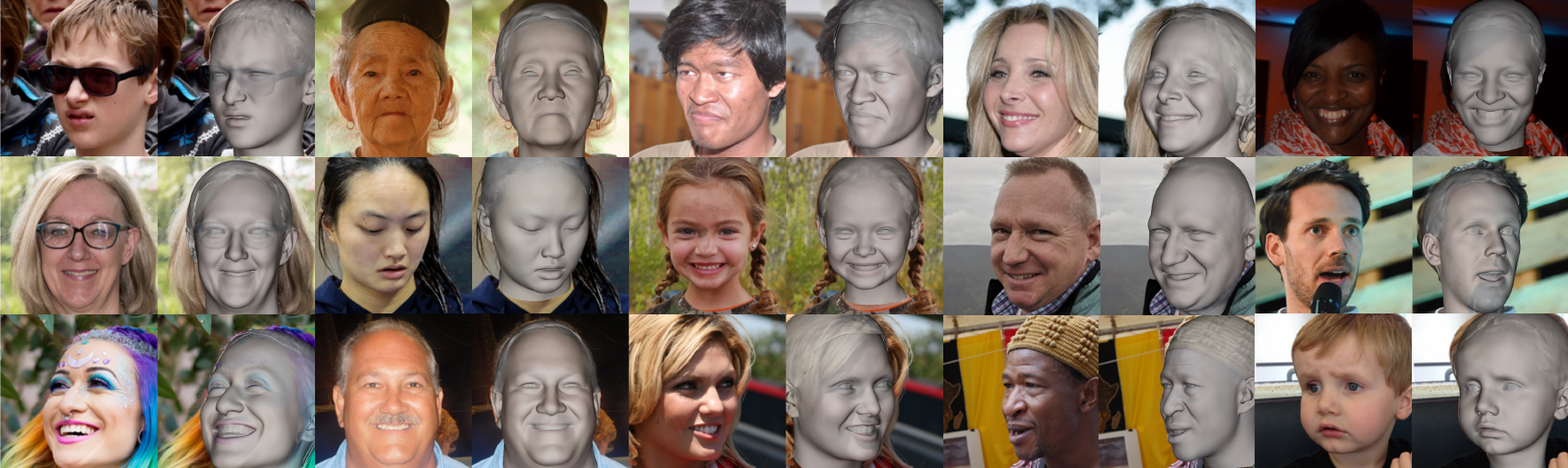}
    }
     \vspace{-0.3cm}
    \captionof{figure}{
    \textbf{\OURS} is a feed-forward network that predicts NPHM~\cite{giebenhain2023nphm} latent codes from a single image. The latent codes can be further optimized at test-time to obtain more detailed 3D reconstructions.
    Here, we show mesh overlays showcasing well-aligned fittings of diverse head shapes and expressions under strong lighting conditions and occlusions. \urlstyle{same}Website: {\url{https://simongiebenhain.github.io/Pix2NPHM/}}
    }
  \label{fig:teaser}
      \vspace{-0.5cm}

 \end{center}
}]

\maketitle
\let\thefootnote\relax\footnotetext{$^*$Providing contracted services for Toyota \\}\par

\vspace{-1.0cm}
\begin{abstract}

      \vspace{-0.5cm}

Neural Parametric Head Models (NPHMs) are a recent advancement over mesh-based 3d morphable models (3DMMs) to facilitate high-fidelity geometric detail.
However, fitting NPHMs to visual inputs is notoriously challenging due to the expressive nature of their underlying latent space.
To this end, we propose {\OURS}, a vision transformer (ViT) network that directly regresses NPHM parameters, given a single image as input. 
Compared to existing approaches, the neural parametric space allows our method to reconstruct more recognizable facial geometry and accurate facial expressions.
For broad generalization, we exploit domain-specific ViTs as backbones, which are pretrained on geometric prediction tasks.
We train {\OURS} on a mixture of 3D data, including a total of over 100K NPHM registrations that enable direct supervision in SDF space, and large-scale 2D video datasets, for which normal estimates serve as pseudo ground truth geometry.
\OURS not only allows for 3D reconstructions at interactive frame rates, it is also possible to improve geometric fidelity by a subsequent inference-time optimization against estimated surface normals and canonical point maps. 
As a result, we achieve unprecedented face reconstruction quality that can run at scale on in-the-wild data.

\end{abstract}    
\section{Introduction}
\label{sec:intro}
Reconstructing faces in 3D, tracking facial movements, and ultimately extracting animation signals for virtual avatars are
fundamental problems in many domains such as the computer games and movie industry, telecommunication, and AR/VR.
Arguably the most relevant sub-task is 3D face reconstruction from a single image due to the vast availability of image collections as well as straight-forward extensions to sequential tracking.

In order to solve the underconstrained reconstruction problem, 3d morphable models (3DMMs)~\cite{blanz1999morphable} have evolved as industry and research standard due to their concise low-dimensional parametric representation, which lead to a plethora of algorithms build on top of 3DMMs.
With the advancement of deep-learning methods, photometric tracking~\cite{thies2016face2face} approaches, have been augmented with additional priors, such as facial landmark detection, or direct 3DMM parameter regression from RGB signal~\cite{Sanyal2019now,feng2021learning_deca,danvevcek2022emoca,zielonka2022towards,zhang2023accurate}. 
Recently, additional priors, such as dense landmarks ~\cite{wood2022denselandmarks,taubner2024flowface} and surface normals~\cite{giebenhain2025pixel3dmm} have further improved reconstructions. 
Due to such methods that enable fitting in even the most challenging scenarios, 3DMMs have become an essential component of photo-realistic avatars~\cite{qian2024gaussianavatars,giebenhain2024npga}, %
generalized avatars~\cite{auth_vol_ava,li2024uravatar,Kirschstein_2025_ICCV_avat3r}, and even controllable generative diffusion models for faces~\cite{ding2023diffusionrig,kirschstein2024diffusionavatars,prinzler2024joker,taubner2024cap4d,taubner2025mvp4d}.

While 3DMMs have achieved great success in these domains, we argue that their concise parametric representation comes at the cost of geometric expressiveness -- i.e., modern 3DMMs, such as FLAME~\cite{FLAME}, are unable to model high-fidelity geometric detail.
Therefore, a more recent line of work has developed neural parametric head models (NPHMs)~\cite{yenamandra2021i3dmm,zheng2022imface,giebenhain2023nphm,giebenhain2024mononphm} for increased representational capacity, as shown in \cref{fig:motivation}.
This increased model capacity, however, makes image-based reconstruction challenging due its expressive parameter space.
MonoNPHM~\cite{giebenhain2024mononphm} has attempted to reconstruct NPHM parameters from a single image. 
However, their purely photometric fitting approach remained slow and brittle in real-world applications %
To this end, we propose a robust and high-fidelity fitting frame-work, yielding a first-class tool for face reconstruction and tracking based on NPHM~\cite{giebenhain2023nphm,giebenhain2024mononphm}.

Our approach addresses the two main challenges of neural parametric model fitting: underconstrained optimization and  reconstruction speed.
This is achieved by tailoring a transformer-based feed-forward predictor for NPHM parameters from a single image.
As a highly data-driven approach, large-scale high-quality training data is essential. To this end, we curated a large collection of publicly available 3D face datasets and fitted MonoNPHM against it, resulting in a total of 102K registrations, which will be shared with the research community. 
Despite these efforts, we find that training on large-scale 2D video datasets using a self-supervised geometric loss based on estimated surface normals~\cite{giebenhain2025pixel3dmm} further improves generalization.
Furthermore, we observe strong generalization improvements by replacing a generic DINOv2~\cite{oquab2023dinov2} backbone with a ViT~\cite{dosovitskiy2020vit} pre-trained on per-pixel geometric prediction tasks, such as surface normal or canonical point map regression.
Our feed-forward estimator produces state-of-the-art (SotA) results, and renders \OURS a first-class choice for 3D face reconstruction, due to its fidelity, ease of use, and robustness to diverse input scenarios, as showcased in \cref{fig:teaser}. 
Moreover, we show that our results can be refined by a few optimization steps at inference time against estimated surface normals to increase fidelity, as shown in  \cref{fig:motivation}.
Together, these insights improve 3D reconstruction on the NeRSemble single-image face reconstruction (SVFR)~\cite{giebenhain2025pixel3dmm} benchmark by 21\%, and neutral reconstruction improves by 6\% compared to the best public method on the NoW~\cite{Sanyal2019now} benchmark.
To summarize, our main contributes are as follows:
\begin{itemize}
    \item {\OURS} is the first feed-forward regressor for NPHM parameters, which, for the first time, enables robust, accurate and fast NPHM reconstructions from a single image.
    \item We curate a large mixture of high-quality 3D datasets, with over 100K NPHM registrations.
    \item For training on 2D data we formulate a novel self-supervised loss using estimated surface normals.
    \item Replacing DINOv2 with geometrically pre-trained, face-specific ViTs improves generalization.
\end{itemize}
\newcolumntype{Y}{>{\centering\arraybackslash}X}
\newcolumntype{P}[1]{>{\centering\arraybackslash}p{#1}}
\begin{figure}[t!]
    \centering
    \includegraphics[width=0.99\linewidth]{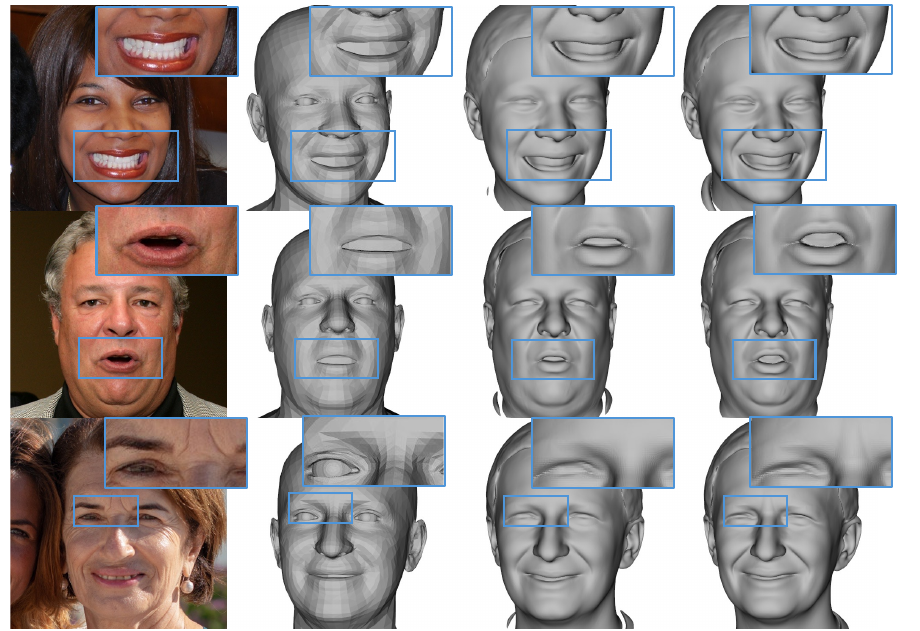}
    \setlength{\tabcolsep}{0pt}
    
\begin{tabularx}{\linewidth}{P{0.23\linewidth}P{0.25\linewidth}YYY}
        \small{Input} & 
        \small{ffwd. FLAME~} & 
        \small{~ffwd. NPHM} & 
        \small{Ours} 
    \end{tabularx}
    \caption{
    \textbf{Motivation:} Single-image 3DMM regressors are limited by their underlying 3DMM. More detailed reconstructions can be obtained by replacing FLAME~\cite{FLAME} with NPHM~\cite{giebenhain2023nphm}, and running inference-time optimization further increase fidelity (see right).
    }
    \label{fig:motivation}
\end{figure}

\section{Related Work}
\paragraph{Optimization-Based Methods}
Early mesh-based 3D Morphable Models (3DMMs)~\cite{blanz1999morphable, bfm09, FLAME} represent facial shape and appearance via linear PCA spaces. Fitting these models to images or videos is typically approached through iterative optimization~\cite{blanz1999morphable, thies2016face2face, zielonka2022towards, grassal2022neural, qian2024gaussianavatars}, usually by minimizing photometric error between renderings and the input. Because photometric losses are under-constrained and sensitive to illumination and noise, many works incorporate geometric priors such as sparse landmarks or semantic segmentations~\cite{thies2016face2face, zielonka2022towards, grassal2022neural, qian2024gaussianavatars} for stability.
More recent methods leverage dense geometric cues to further improve robustness and accuracy, such as dense landmarks~\cite{wood2022denselandmarks}, UV flow~\cite{taubner2024flowface}, or a combination of UV maps and surface normals~\cite{giebenhain2025pixel3dmm}.

\begin{figure*}[htb!]
    \centering
    \includegraphics[width=0.99\textwidth]{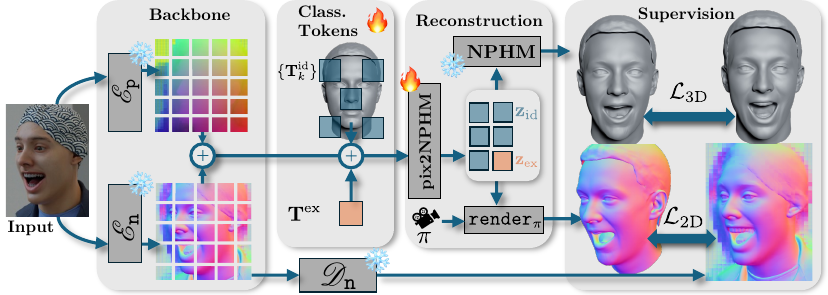}

    \caption{
    \textbf{Method Overview:} 
    We use pretrained ViTs, $\mathscr{E}_{\texttt{n}}$ and $\mathscr{E}_{\texttt{p}}$, as backbone, which encode the input into a token sequence.
    The resulting sequence is concatenated with learnable classifier token $\{\mathbf{T}^{\text{id}}_k\}$ and $\mathbf{T}^{\text{ex}}$, which are decoded into NPHM identity ($\mathbf{z}_{\text{id}}$) and expression~($\mathbf{z}_{\text{ex}}$) parameters using transformer network.
    We train using a 3D SDF loss, and a normal rendering loss against pseudo g.t. normals.
    }
    \label{fig:method}

\end{figure*}

\paragraph{Regression-Based Methods}
Regression-based approaches train deep networks to directly predict 3DMM parameters, avoiding iterative optimization and enabling real-time inference. Fully supervised approaches~\cite{richardson20163d, richardson2017learning, tuan2017regressing, trần2018extreme, guo20203ddfav2, zielonka2022towards} rely on registered 3D scans or synthetic data, but are limited by the scarcity of high-quality annotations.
Self-supervised methods learn from in-the-wild images by combining differentiable rendering with photometric constraints~\cite{tewari2017mofa, Sanyal2019now, feng2021learning_deca, danvevcek2022emoca, filntisis2022visual, zhang2023accurate, SMIRK:CVPR:2024, Schoneveld_2025_ICCV}. EMOCA~\cite{danvevcek2022emoca} extends this by incorporating emotion-aware supervision for more expressive results, and SPECTRE \cite{filntisis2022visual} integrated lipreading-based cues to better capture mouth movements and speech-related expressions. TokenFace~\cite{zhang2023accurate} combines 3D and 2D supervision in a ViT-based framework.
Recent work highlights the role of improved differentiable rendering: SMIRK~\cite{SMIRK:CVPR:2024} employs a neural renderer, while SHeaP~\cite{Schoneveld_2025_ICCV} uses 2D Gaussian splats~\cite{Huang2DGS2024}.\\
Our method follows this self-supervised line but replaces photometric cues with surface-normal supervision and builds upon MonoNPHM rather than FLAME.

\paragraph{Neural 3DMMs}
Classical 3DMMs~\cite{blanz1999morphable, bfm09, FLAME} rely on fixed topology and linear PCA spaces, limiting detail and expressiveness. Neural 3DMMs address these issues by using implicit neural representations such as SDFs~\cite{yenamandra2021i3dmm, zheng2022imface, zheng2023imface++, palafox2021npms, giebenhain2023nphm} or occupancy grids~\cite{chen2022gdna}.
NPHM~\cite{giebenhain2023nphm} models head shape and appearance via local SDF experts anchored at semantic keypoints, and uses a deformation field for expressions. MonoNPHM~\cite{giebenhain2024mononphm} proposed the first monocular fitting method for NPHM through volumetric SDF rendering and canonical appearance modeling.
Other works explore photorealistic neural head models using NeRFs~\cite{mildenhall2021nerf} and 3D Gaussian splats~\cite{kerbl3Dgaussians}, such as \cite{mildenhall2021nerf, kerbl3Dgaussians, morf, buehler2024cafca, xu2024gphm, zheng2024headgap}. \\
Compared to FLAME, NPHM’s manifold has proven valuable for downstream tasks, including realistic avatars~\cite{kirschstein2024diffusionavatars, giebenhain2024npga} and high-fidelity audio-driven geometry~\cite{aneja2023facetalk}.
Motivated by this, we develop a fast, robust, and accurate reconstruction approach that makes MonoNPHM practical for broader downstream use.

\section{Pix2NPHM}
Given a single input image $I$ as input, it is our goal to estimate identity parameters $\mathbf{z}_{\text{id}}$ and expression parameters $\mathbf{z}_{\textbf{ex}}$ from which the 3D head geometry can be recovered using MonoNPHM~\cite{giebenhain2024mononphm} as decoder.
After providing background information on NPHMs in \cref{sec:background}, we begin by describing our face-specific pre-training strategy to obtain our image encoding backbone in \cref{sec:pretraining}.
Next, in \cref{sec:formulation} we introduce {\OURS}, the first method for feed-forward NPHM parameter regression based on a single image. \cref{fig:method} provides an overview of our network architecture, and training strategy, which is further described in \cref{sec:training}. 
Finally, \cref{sec:inference_time} describes our inference-time optimization, which can further improve our reconstruction fidelity, as showcased in \cref{fig:motivation}.

\subsection{Background: Neural 3DMMs}
\label{sec:background}

In this paper we adopt the neural 3DMM formulation introduced by MonoNPHM~\cite{giebenhain2024mononphm}. It defines a neural field
\begin{equation}
    \mathscr{F}_{\text{NPHM}} : (x, \mathbf{z}_{\text{id}}, \mathbf{z}_{\text{ex}} ) \mapsto \text{SDF}(x),
    \label{eq:nphm}
\end{equation}
which predicts SDF-values for the 3D head geometry, given points $x\in~\mathbb{R}^3$, and is conditioned on disentangled identity and expression latent codes. 
Internally, $\mathscr{F}_{\text{NPHM}}$ consists of a backward deformation field, conditioned on $\mathbf{z}_{\text{ex}}$ and $\mathbf{z}_{\text{id}}$ and a canonical SDF conditioned on $\mathbf{z}_{\text{id}}$.
A mesh can be extracting using marching cubes~\cite{marchingcubes}, once latent codes have been obtained. 
The remainder of the paper will focus on the inference of ideal latent codes from a single image.
Throughout all our experiments we use the public checkpoint from MonoNPHM~\cite{giebenhain2024mononphm}.

\subsection{Robust Encoder via Geometric Pre-Training}
\label{sec:pretraining}

We tackle the inherent ambiguity of the single-image reconstruction task using a heavily data-driven approach. 
Our method starts with pre-training a ViT-based backbone on a per-pixel geometric reconstruction task. We follow the training strategy and encoder-decoder architecture 
\begin{eqnarray}
    \mathscr{E}_{\texttt{geo}}&: \mathbb{R}^{H\times W \times 3} \rightarrow \mathbb{R}^{L \times D} \\
    \mathscr{D}_{\texttt{geo}}&:  \mathbb{R}^{L \times D} \rightarrow \mathbb{R}^{H\times W \times 3} 
\end{eqnarray}
from Pixel3DMM~\cite{giebenhain2025pixel3dmm}, where $\texttt{geo} \in \{ \texttt{n}, \texttt{p} \}$ represent surface normal and canonical point cloud estimators, respectively. 
While point cloud prediction has to be handled using relative coordiante systems for arbitrary scenes, such as in DUSt3R~\cite{wang2024dust3r} and VGGT~\cite{wang2025vggt}, we exploit the possibility to define a unique coordinate system for 3D heads.
Once pre-training is completed, we discard  $\mathscr{D}_{\texttt{geo}}$, and use the face-specific geometrically aware encoders $\mathscr{E}_{\texttt{geo}}$ as backbone for our regression network. Later we will show that this face-specific pre-training significantly outperforms DINOv2 image encodings.

\subsection{Feed-Forward NPHM Parameter Prediction}
\label{sec:formulation}
Following the recent success of transformer networks, we employ learnable classifier tokens, similar to ViT~\cite{dosovitskiy2020vit} and TokenFace~\cite{zhang2023accurate}, which read out relevant identity and expression information after several transformer layers.
To this end we concatenate our encoded token sequence $[\mathscr{E}_{\texttt{n}}(I), \mathscr{E}_{\texttt{p}}(I)]$ with learnable classifier tokens 
\begin{equation}
    \textbf{T}_{\text{CLS}} = \left[ \textbf{T}^{\text{ex}}, \{\textbf{T}^{\text{id}}_k\}_{k=1}^{66} \right]
\end{equation}
for expression and identity. Note that $\textbf{T}^{\text{id}}$ is composed of 65 tokens, one for each local identity codes from {MonoNPHM}~\cite{giebenhain2024mononphm}, and one for the global identity part. 
Finally, our regressor
\begin{equation}
    \text{pix2NPHM} : 
    ([\mathscr{E}_{\texttt{n}}(I), \mathscr{E}_{\texttt{p}}(I)], \mathbf{T}_{CLS}) \mapsto \left( \mathbf{z}_{\text{id}}, \mathbf{z}_{\text{ex}} \right) 
\end{equation}
 consists of several transformer layers, and MLP prediction heads which map the resulting classification tokes tokens $\mathbf{T}^\prime_{\text{CLS}}$ from the transformer to $\mathbf{z}_{\text{id}}$ and $\mathbf{z}_{\text{ex}}$, respectively.
Finally, the predicted latent codes can be decoded into a 3D shape using the MonoNPHM model, which  represents the 3D head geometry implicitly by conditioning an SDF, see \cref{eq:nphm}

\subsection{Training Strategy}
\label{sec:training}

\subsubsection{3D supervision in SDF Space}
On 3D datasets we aim to formulate the supervision signal as unambiguous as possible, and exploit the fact that ground truth NPHM codes $(\mathbf{z}_{\text{id}}^{\text{gt}}, \mathbf{z}_{\text{ex}}^{\text{gt}})$ can be obtained as a pre-processing step using our registration procedure, as described in \cref{sec:data_registration}. 
Given an image $I$, we predict NPHM latents $\text{pix2NPHM}(I, \mathbf{T}_{CLS})=\left(\hat{\mathbf{z}}_{\text{id}}, \hat{\mathbf{z}}_{\text{ex}}\right)$, and compute the loss 
\begin{equation}
    \mathcal{L}_{\text{3D}}\!=\!\sum_{x\in \mathcal{X}}\Vert 
    \text{NPHM}(x; \hat{\mathbf{z}}_{\text{id}}, \hat{\mathbf{z}}_{\text{ex}}
    )\!-\!\text{NPHM}(x; \mathbf{z}_{\text{id}}^{\text{gt}}, \mathbf{z}_{\text{ex}}^{\text{gt}})
    \Vert_1,
\end{equation}
between the induced SDFs, where $\mathcal{X}$ is a point cloud randomly sampled near the 3D surface. Note, that directly supervising $\Vert \hat{\mathbf{z}} - \mathbf{z}\Vert$ did not lead to training convergence.
\subsubsection{2D Self-Supervision: Adding Diversity}
Since 3D datasets barely cover all relevant modes of the valid input distribution, we add data diversity by training on large scale 2D video datasets using estimated surface normals $I^{\texttt{n}} = \mathscr{D}_{\texttt{n}}\left(\mathscr{E}_{\texttt{n}}\left(I\right)\right)$ as pseudo ground truth.
To this end, we modify the  NeuS-based~\cite{wang2021neus} rendering approach from MonoNPHM~\cite{giebenhain2024mononphm} to render surface normals, and supervise via the cosine similarity
\begin{equation}
    \mathcal{L}_{\text{2D}}^{\texttt{n}} = \sum_{\mathbf{p} \in \mathcal{P}} \langle\texttt{render}_{\pi}\left(\text{NPHM}; \hat{\mathbf{z}}_{\text{id}}, \hat{\mathbf{z}}_{\text{ex}}\right)_{\mathbf{p}}, 
    I^{\texttt{n}}_{\mathbf{p}} \rangle
    \label{eq:render}
\end{equation}
between rendered and estimated normals, where camera parameters $\pi$ are provided by our data registration. Note that due to the memory-intensive nature of MLP-based volumetric rendering we only supervise a random subset of pixels $\mathbf{p} \in \mathcal{P}$ sampled in the facial area.
Overall, we find this supervision to provide much more meaningful and stable gradients, compared to computing a photometric loss using spherical harmonics, which is the most established loss function in existing FLAME-based regressors~\cite{feng2021learning_deca,Schoneveld_2025_ICCV}.

\subsubsection{Full Training Objective}
Overall, we train our feed-forward predictor using 
\begin{equation}
    \mathcal{L}_{\text{total}} = \lambda_{\text{3D}}\mathcal{L}_{\text{3D}} + \lambda_{\text{2D}}\mathcal{L}_{\text{3D}} + \lambda_{\text{reg}}\mathcal{R}
\end{equation}
as our complete training objective, which combines 3D and 2D losses with regularization terms,
where $\lambda_{\text{3D}}$ is set to $0$ for 2D video datasets. 
Our regularization term
\begin{equation}
    \mathcal{R}(\mathbf{z}_{\text{id}}, \mathbf{z}_{\text{ex}}) = \lambda^{\mathcal{R}}_{\text{id}}\Vert \mathbf{z}_{\text{id}}\Vert_2
+
\lambda^{\mathcal{R}}_{\text{ex}}\Vert \mathbf{z}_{\text{ex}}\Vert_2
\end{equation}
simply punishes the norm of the predicted latents.

\subsection{Test-Time Optimization: Increasing Fidelity}
\label{sec:inference_time}

For even more accurate 3D reconstructions our feed-forward estimates can be naturally combined with inference-time optimization against per-pixel geometric predictions, similar to Pixel3DMM~\cite{giebenhain2025pixel3dmm}.
We optimize for
\begin{equation}
   \argmin_{\mathbf{z}_{\text{id}}, \mathbf{z}_{\text{ex}}, \pi} \lambda_{\texttt{n}}\mathcal{L}_{\text{2D}}^{n} + \lambda_{\texttt{p}}\mathcal{L}_{\text{2D}}^{p} + \lambda_{\text{reg}}\mathcal{R}\left(\mathbf{z}_{\text{id}}\!-\!\hat{\mathbf{z}}_{\text{id}}, \mathbf{z}_{\text{ex}}\!-\!\hat{\mathbf{z}}_{\text{id}}\right),
\end{equation}
where our feed-forward predictions serve as initialization and regularization target, and $\mathcal{L}^{\texttt{p}}_{\text{2D}}$ is defined similar to \cref{eq:render} but with an $L_1$-Loss.
Note that we first need to estimate camera parameters $\pi$, which includes the head pose, using dense landmark predictions from Pixel3DMM, which are fine-tuned later alongside $\mathbf{z}_{\text{id}}$ and $\mathbf{z}_{\text{ex}}$.

\subsection{High-Quality Training Dataset Curation}
\label{sec:data_registration}

As a data-driven approach, a core requirement for success is sufficient high-quality training data. Therefore, we spend a considerable amount of effort on curating 3D and 2D training datasets. We briefly describe our approach for different data types below, and provide an overview of the used training data in table~\cref{tab:training_data}.
\begin{table}[]
\resizebox{\columnwidth}{!}{

\begin{tabular}{@{}lcccc@{}}
\toprule
\multicolumn{1}{l}{} & \#IDs  & \small \#Expr./\#Views     & \small \#Images & \small \#3D shapes \\ \midrule
\small NPHM~\cite{giebenhain2023nphm}        & 450            & 23                   / 40                   & 414K              & 10K                 \\
\small FaceScape~\cite{zhu2023facescape}   & 300            & 20                   / 50             & 300K              & 6K                  \\
\small DAViD~\cite{saleh2025david}       & 65K          & 1                    / 1                    & 65K               & 65K                 \\
\small MimicMe~\cite{thanos2022_mimicme}     & 2000           & 10                   / 5                    & 100K              & 20K                 \\
\small LYHM~\cite{dai2020statistical_lyhm}        & 1200           & 1                    / 2                    & 2.4K              & 1.2K                \\
\small Videos~\cite{cui2024hallo3,zhu2022celebvhq,yu2022celebvtext}& \textless{}50K & 5                    / 1                    & 250K              & 0                   \\ \midrule
\small Total 3D   & 69K            & -/-   & 880K              & 102K                \\ \midrule
\small Total       & 119K           & -/-  & 1.13M             & 102K \\ \bottomrule

\end{tabular}
}
\caption{\textbf{Training Data:}
We list number of identites, facial expressions, camera view-points, total images and total 3D shapes.%
}
\label{tab:training_data}
\end{table}

\paragraph{3D datasets}
For 3D datasets, we estimate the canonical coordinate frame using FLAME registration similar to \cite{giebenhain2023nphm}.
After transforming the ground truth 3D shapes into a unified coordinate system, we sample points $x\in\mathcal{X}$ on the ground truth surface and optimize for NPHM parameters 
\begin{equation}
    \argmin_{\mathbf{z}_{\text{id}}, \mathbf{z}_{\text{ex}}} \sum_{x\in \mathcal{X}}\Vert \text{NPHM}(x;\mathbf{z}_{\text{id}}, \mathbf{z}_{\text{ex}})\Vert_1 + \lambda_{\text{reg}}\mathcal{R}(\mathbf{z}_{\text{id}}, \mathbf{z}_{\text{ex}})
\end{equation}
\paragraph{2D datasets}

For large-scale, in-the-wild 2D datasets, obtaining reliable and 3D accurate NPHM fittings was a previously unsolved problem.
Therefore, we simply estimate camera poses using the video tracker from Pixel3DMM~\cite{giebenhain2025pixel3dmm}, and purely rely on our rendering loss $\mathcal{L}_{\text{2D}}^{\texttt{n}}$.

\paragraph{Data Cleaning}

To ensure high-quality training data, we devise a simple outlier filtering strategy, for both, 3D NPHM fittings and 2D FLAME video fittings. We define outlier thresholds, by analyzing the histograms of the norms of certain attributes, such as shape, expressions, neck and jaw parameters.

\newcolumntype{Y}{>{\centering\arraybackslash}X}
\newcolumntype{P}[1]{>{\centering\arraybackslash}p{#1}}
\begin{figure*}[htb!]
    \centering
    \includegraphics[width=0.99\textwidth]{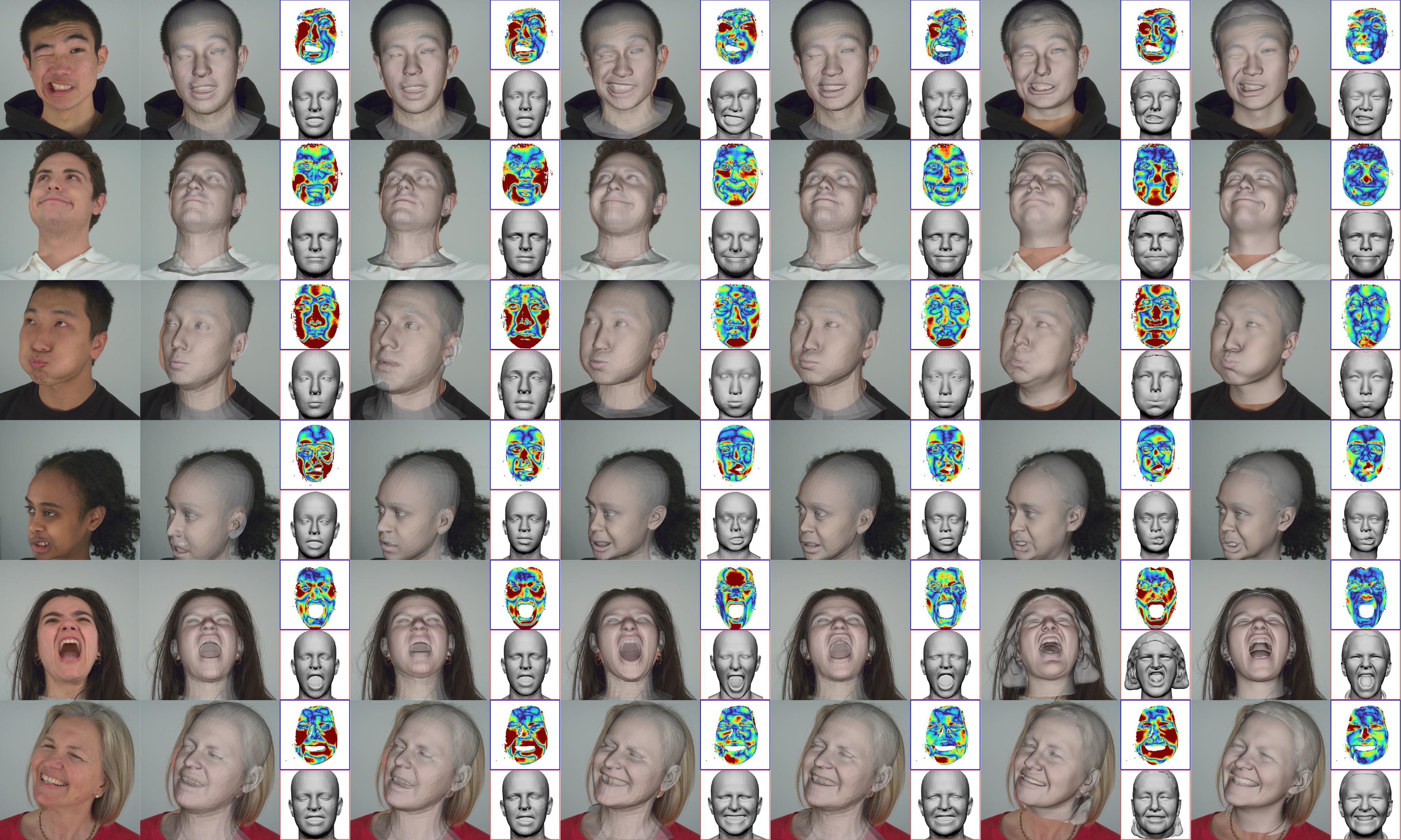}
    \setlength{\tabcolsep}{0pt}
    
    \begin{tabularx}{\textwidth}{P{0.12\linewidth}P{0.12\linewidth}YYYYY}
        \small{Input} & 
        \small{DECA}~\cite{feng2021learning_deca} & 
        \small{TokenFace}~\cite{zhang2023accurate} & 
        \small{FlowFace}~\cite{taubner2024flowface} &
        \small{Pixel3DMM}~\cite{giebenhain2025pixel3dmm} &
        \small{MonoNPHM}~\cite{giebenhain2024mononphm} &
        \small{Ours}
    \end{tabularx}
    \caption{
    \textbf{Posed Reconstruction:} We show overlays of the reconstructed meshes to judge the reconstruction alignment. Insets with a blue border depict $L_2$-Chamfer distance as an error map, rendered from a frontal camera. Red insets show the reconstructed mesh from the same camera. All our figures are best viewed digitally and zoomed-in.
    }
    \label{fig:main_results_posed}
\end{figure*}

\section{Experimental Results}
\label{sec:experiments}

To verify our proposed method experimentally, we compare against recent SotA methods on the NeRSemble single-view face reconstruction (SVFR) benchmark~\cite{giebenhain2025pixel3dmm} (see \cref{sec:results_svfr}), and on the established Now benchmark~\cite{Sanyal2019now} (see \cref{sec:results_now}).
In \cref{sec:ablations} we conduct thorough ablations experiments to quantify the significance of our individual technical contributions.
We highly encourage the reviewers to watch our supplementary video for more results.  We will release our training data, model and full code base for research purposes.

 \subsection{Implementation Details} 
We implement our pipeline using PyTorch, and train our regression model using the Adam~\cite{adam} optimizer, a batch size of 32 and learning rate of $1e^{-4}$ on a single A100-80GB GPU, which takes 4 days until convergence. We set $\lambda_{\text{3D}}{=}10.0$, $\lambda_{\text{2D}}{=}1.0$ and $\lambda_{\text{reg}}{=}1e^{-4}$. 
We use $8$ transformer layers, each consisting of pre-norm LayerNorm, self-attention with $8$ heads, a 2-layer MLP with GeLU activation, and a width of $1024$.\\
We pre-train $\mathscr{E}_{n}$ and $\mathscr{E}_{p}$ separately for 3 days on 2 A6000 GPUs, and follow the network architecture and training strategy fom Pixel3DMM~\cite{giebenhain2025pixel3dmm}.\\
During inference our complete feed-forward estimator runs at $8$fps on an RTX3080-10GB. For optimization we perform 100 steps, which takes $85s$ on an RTX3080. For more details on the optimization procedure, we refer to our supplementary material.

To register our training dataset, consisting of 102K 3D shapes, with the NPHM model, we invested roughly $2.500$ GPU-hours. We will release our training data to the community to drive future research.

 \begin{table}[]
\centering
\setlength{\tabcolsep}{1.8pt}
\small
\begin{tabular}{@{}lrrrrrrr}
\toprule
    \multicolumn{1}{c}{\multirow{2}{*}{Method}}     & \multicolumn{3}{c}{Neutral}              && \multicolumn{3}{c}{Posed}                 \\
          \cmidrule(l){2-4}  \cmidrule(l){6-8}
         & L1\scriptsize{$\downarrow$}       & L2\scriptsize{$\downarrow$}        & NC\scriptsize{$\uparrow$}             &&  L1\scriptsize{$\downarrow$}       & L2\scriptsize{$\downarrow$}        & NC\scriptsize{$\uparrow$}                \\ \midrule
MICA~\cite{zielonka2022towards}     
    & 1.68 & 1.14 & 0.883  &&  -         &   -      &   -         \\
TokenFace~\cite{zhang2023accurate} 
    & - & - & -  && 2.62 & 1.78 & 0.865  \\
DECA~\cite{feng2021learning_deca}     
    & 2.07         &  1.40         &    0.876             &&  2.38         &   1.61        &   0.870          \\
EMOCAv2\cite{danvevcek2022emoca}  
    &    2.21      &   1.49        &    0.873         &&   2.63        &    1.78       &     0.860        \\
SHeaP~\cite{Schoneveld_2025_ICCV} & 1.86 &  1.26 &  0.882
 && 2.08 & 1.41 &  0.876 \\
    
    Ours (ffwd. only) & 1.57 &  1.06 &  0.896    && 1.55 &  1.05 &  0.894 \\ \hline
Metr. Tracker\cite{zielonka2022towards} 
    & - & - & -  &&    2.03 &  1.37 &  0.878      \\

FlowFace~\cite{taubner2024flowface} 
    & 1.93 & 1.31 &  0.878  && 1.96   & 1.33   & 0.879      \\
Pixel3DMM~\cite{giebenhain2025pixel3dmm}     
    & 1.66  & 1.12  & 0.883   && 1.66 & 1.11   & 0.884  \\
MonoNPHM~\cite{giebenhain2024mononphm}     
    & 2.32  & 1.56  & 0.878   && 2.50 & 1.68   & 0.870  \\
Ours     
    & \textbf{1.54} &  \textbf{1.04} &  \textbf{0.897}   && \textbf{1.37} &  \textbf{0.92} &\textbf{0.897} \\
    \bottomrule
\end{tabular}
\caption{\textbf{NeRSemble-SVFR Benchmark~\cite{giebenhain2025pixel3dmm}:} Single image \emph{posed} and \emph{neutral} 3D face reconstruction. Methods above the line represent feed-forward networks, methods below require optimization.
}
\label{tab:main_results}
\vspace{-0.3cm}
\end{table}

\begin{table}[]
\centering
\begin{tabular}{@{}lccc|c@{}}
\toprule
\multicolumn{1}{c}{\multirow{2}{*}{Method}} & \multicolumn{3}{c|}{Error (mm)}                                                  & \multirow{2}{*}{Avail.} \\ \cmidrule(lr){2-4}
\multicolumn{1}{c}{}                      & \multicolumn{1}{l}{Median} & \multicolumn{1}{l}{Mean} & \multicolumn{1}{l|}{Std} &                        \\ \midrule
DECA~\cite{feng2021learning_deca}                                       & 1.09                       & 1.38                     & 1.18                     & \checkmark                      \\
MICA~\cite{zielonka2022towards}                                    & 0.90                       & 1.11                     & 0.92                     & \checkmark                      \\
SHeaP~\cite{Schoneveld_2025_ICCV}                                      & 0.95                       & 1.18                     & 0.99                     & \checkmark                      \\
TokenFace~\cite{zhang2023accurate}                                  & 0.76                       & 0.95                     & 0.82                     & \xmark                      \\

Ours (ffwd. only)                                      & 0.83                       & 1.03                     & 0.88                     & \checkmark$^*$                     \\ \hline
FlowFace~\cite{taubner2024flowface}                                   & 0.87                       & 1.07                     & 0.88                     & \xmark                      \\
Pixel3DMM~\cite{giebenhain2025pixel3dmm}                                  & 0.87                       & 1.07                     & 0.89                     & \checkmark                      \\
Ours                                       & 0.81                       & 1.01                     & 0.85                     & \checkmark$^*$                     \\ \bottomrule
\end{tabular}
\label{tab:results_now}
\caption{\textbf{NoW Benchmark}~\cite{Sanyal2019now}:
Single-image neutral 3D face reconstruction. Methods above the line are feed-forward networks. 
}

\label{tab:results_now}
\end{table}

\newcolumntype{Y}{>{\centering\arraybackslash}X}
\newcolumntype{P}[1]{>{\centering\arraybackslash}p{#1}}
\begin{figure}[htb!]
    \centering
    \vspace{-2mm}
    \includegraphics[width=0.99\linewidth]{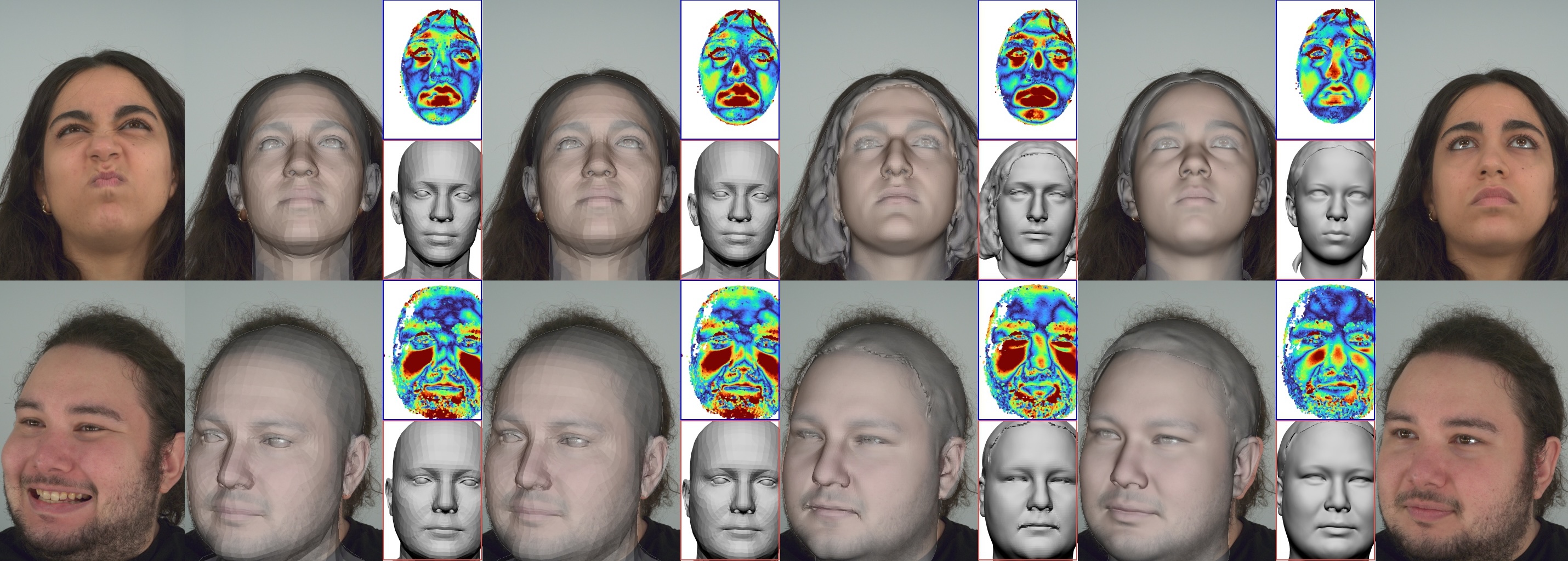}
    \setlength{\tabcolsep}{0pt}
    \vspace{-2mm}
    \begin{tabularx}{\linewidth}{P{0.15\linewidth}P{0.15\linewidth}P{0.2\linewidth}P{0.2\linewidth}P{0.15\linewidth}P{0.15\linewidth}}
        \scriptsize{Input} & 
        \scriptsize{MICA}~\cite{zielonka2022towards} & 
        \scriptsize{Pixel3DMM}~\cite{giebenhain2025pixel3dmm} &
        \scriptsize{MonoNPHM}~\cite{giebenhain2024mononphm} &
        \scriptsize{Ours} &
        \scriptsize{Reference}
    \end{tabularx}
    \caption{
    \textbf{Neutral Reconstruction, NeRSemble:} Comparison against available SotA methods on top of neutral reference image. 
    }
    \label{fig:main_results_neutral}
\end{figure}
\newcolumntype{Y}{>{\centering\arraybackslash}X}
\newcolumntype{P}[1]{>{\centering\arraybackslash}p{#1}}
\begin{figure}[htb!]
    \centering
    \vspace{-2mm}
    \includegraphics[width=0.99\linewidth]{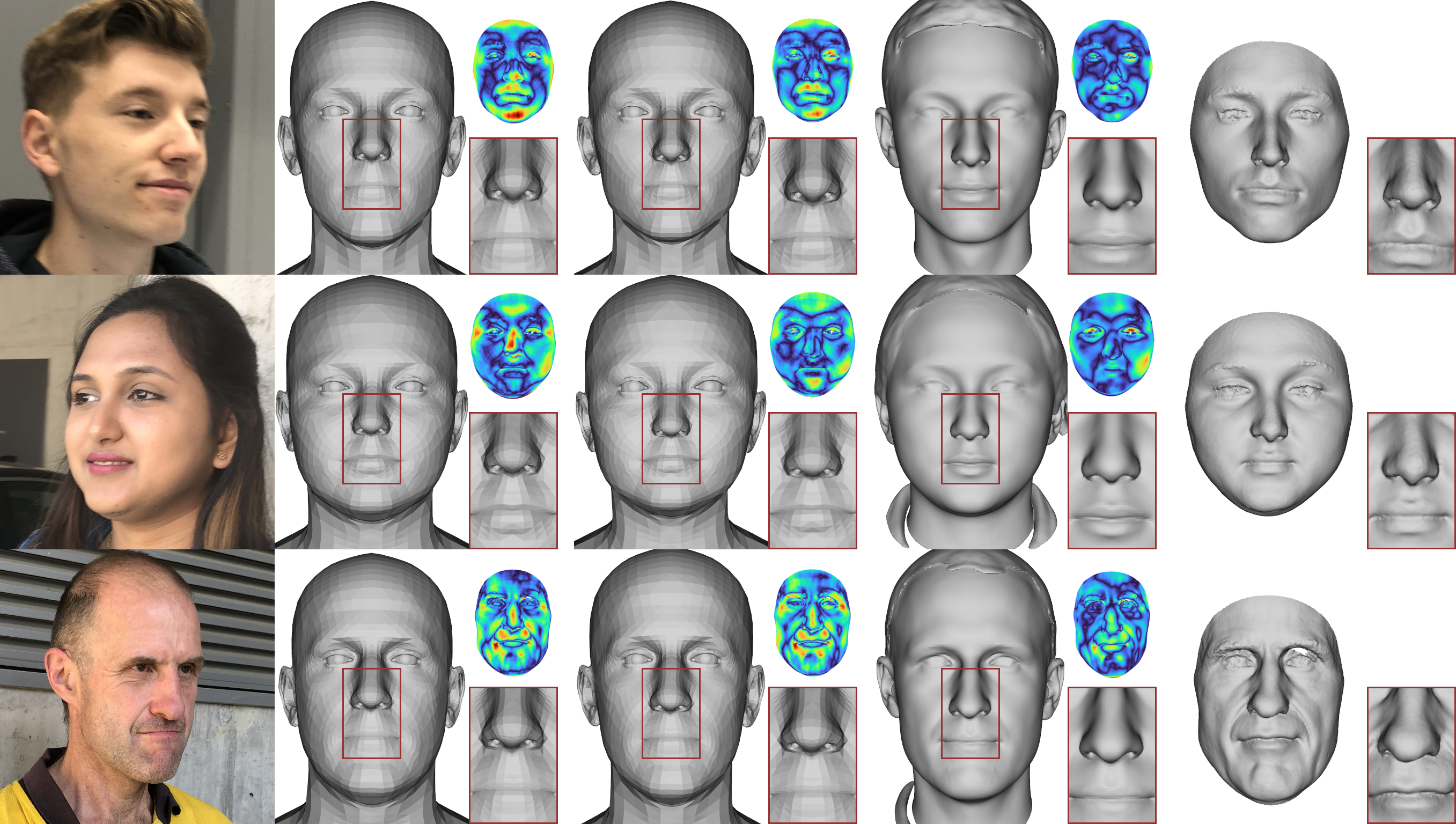}
    \setlength{\tabcolsep}{0pt}
    \vspace{-2mm}
    \begin{tabularx}{\linewidth}{P{0.17\linewidth}P{0.20\linewidth}YYY}
        \scriptsize{Input} & 
        \scriptsize{MICA}~\cite{zielonka2022towards} & 
        \scriptsize{Pixel3DMM}~\cite{giebenhain2025pixel3dmm} &
        \scriptsize{Ours} &
        \scriptsize{Ground Truth}
    \end{tabularx}
    \caption{
    \textbf{Neutral Reconstruction, NoW:} We show frontal mesh renderings in comparison to the ground truth mesh, as well as, error maps and zoom ins.
    }
    \label{fig:main_results_neutral_now}
\end{figure}

 \subsection{Baselines}

 \paragraph{Feed-Forward FLAME Regressors}
The first class of baselines are existing regressors for FLAME~\cite{FLAME} parameters. We select DECA~\cite{feng2021learning_deca}, MICA~\cite{zielonka2022towards}, EMOCAv2~\cite{danvevcek2022emoca}, TokenFace~\cite{zhang2023accurate} and SHeaP~\cite{Schoneveld_2025_ICCV}.

 \paragraph{FLAME Optimization}
 The next class combines feed-forward predictions as a prior with inference-time optimization. As such, this category is closely related to our full approach, but uses a classical 3DMM representation. We select MetricalTracker~\cite{zielonka2022towards}, FlowFace~\cite{taubner2024flowface} and Pixel3DMM~\cite{giebenhain2025pixel3dmm}.

 \paragraph{NPHM Optimization}
 Finally, MonoNPHM~\cite{giebenhain2024mononphm} proposes a photometric fitting pipeline to obtain NPHM parameters. Since we also use the MonoNPHM model to decode latent parameters into 3D shapes, this comparison highlights the added robustness of our approach compared to photometric fitting.

\paragraph{Baseline Availability}
Finally, we point out the poor public availability of recent SotA methods, \ie TokenFace~\cite{zhang2023accurate} and FlowFace~\cite{taubner2024flowface}, two of the top performing methods on NoW, remain unreleased, and haven't been replicated as of yet.

 \subsection{NeRSemble SVFR Benchmark}
 \label{sec:results_svfr}

The recent NeRSemble SVFR benchmark~\cite{giebenhain2025pixel3dmm} features diverse and challenging facial expressions, and is the first to simultaneously allow the evaluation of \emph{posed} and \emph{neutral} face reconstruction. Given a single image as input, the \emph{posed} task measures the 3D reconstruction accuracy, while the \emph{neutral} task measures how well the face under neutral expression can be reconstructed.
We provide quantitative and qualitative results in \cref{tab:main_results} and \cref{fig:main_results_posed,fig:main_results_neutral}, respectively.
We significantly outperform all baselines across all metrics, even without test-time optimization.

 \subsection{NoW Benchmark}
 \label{sec:results_now}

 The NoW benchmark~\cite{Sanyal2019now} can only measure the \emph{neutral} reconstruction task, but, compared to the NeRSemble SVFR benchmark, it covers a wider variety of identities, lighting conditions, hair styles, head accessories and other types of occlusion.
 Our method significantly outperforms all baselines, except for TokenFace~\cite{zhang2023accurate}, which is, however, not publicly available and performs poorly on the NeRSemble SVFR benchmark, as presented in \cref{tab:results_now}.
 Qualitative results from the validation set are presented in \cref{fig:main_results_neutral_now}.

\subsection{AffectNet Benchmark}

Although emotion recognition is not directly measuring any sort of 3D understanding, we further assess the degree to which the semantics of facial expressions are accurately captured by our feedforward network on AffectNet~\cite{mollahosseini2017affectnet}. We extract shape and expression parameters using \OURS{} for all train and test images in the AffectNet dataset (note that this is done in a feedforward-only manner, without per-image optimization). Following EMOCA~\cite{danvevcek2022emoca}, we then fit a 4-layer MLP to the the parameters extracted from the train set, and evaluate on the test set. \OURS{} outperforms all competing methods on this benchmark (Table \ref{tab:affectnet}), showcasing our trained feedforward network's ability to extract detailed expression information.

\begin{table}[t]
\centering
\setlength{\tabcolsep}{4pt} %
\begin{tabular}{lccccc}
\toprule
\multirow{2}*{Model} & \multicolumn{2}{c}{Arousal} & \multicolumn{2}{c}{Valence} & Emo \\
                        \cmidrule(lr){2-3} \cmidrule(lr){4-5}  \cmidrule(lr){6-6}
                    & CCC↑ & RMSE↓ & CCC↑ & RMSE↓ & Acc. \\
\midrule
SMIRK~\cite{SMIRK:CVPR:2024} & 0.560 & 0.288 & 0.681 & 0.313 & 0.653 \\
EMOCA~\cite{danvevcek2022emoca} & 0.577 & 0.282 & 0.70 & 0.307 & 0.676 \\
SHeaP~\cite{Schoneveld_2025_ICCV} & 0.615 & 0.274 & 0.735 & 0.301 & 0.695 \\

Ours-FLAME & 0.553 & 0.289 & 0.702 & 0.305 & 0.660 \\
Ours-NPHM & \textbf{0.621} & \textbf{0.273} & \textbf{0.739} & \textbf{0.291} & \textbf{0.711} \\
\bottomrule
\end{tabular}
\vspace{-2mm}
\caption{
\textbf{Emotion recognition:} We report the concordance correlation coefficient (CCC) and root mean squared error (RMSE) on predicting Valence and Arousal, and, most importantly, we report accuracy in 8-way emotion classification on AffectNet~\cite{mollahosseini2017affectnet}.}
\label{tab:affectnet}
\vspace{-2mm}
\end{table}

\begin{table}[]
\resizebox{\columnwidth}{!}{
\begin{tabular}{@{}l|lllll|l@{}}
\toprule
Method          & 2D & 3D & Input & Opt. & 3DMM  & Error \\ \midrule
3D only       &    & \checkmark  & $\mathscr{E}_{n}+\mathscr{E}_{\texttt{p}}$     &         & NPHM  & 1.074 \\
2D only       & \checkmark  &    & $\mathscr{E}_{n}+\mathscr{E}_{\texttt{p}}$     &         & NPHM  & 1.238 \\ \hline
RGB Input     &    & \checkmark  &$\text{DINO}(I)$   &         & NPHM  & 1.165 \\
Normals Inp. & \checkmark  & \checkmark  & $\text{DINO}(I^{\texttt{n}})$  &         & NPHM  & 1.168 \\
no $\mathscr{E}_{\texttt{p}}$       & \checkmark  & \checkmark  & $\mathscr{E}_{n}$    &         & NPHM  & 1.053 \\ \hline
ffwd NPHM     & \checkmark  & \checkmark  & $\mathscr{E}_{n}+\mathscr{E}_{\texttt{p}}$     &         & NPHM  & 1.016 \\
ffwd. FLAME    & \checkmark  & \checkmark  & $\mathscr{E}_{n}+\mathscr{E}_{\texttt{p}}$     &         & FLAME & 1.073 \\
\hline

opt. only w/o MICA     & \checkmark  & \checkmark  & $\mathscr{E}_{n}+\mathscr{E}_{\texttt{p}}$     & \checkmark       & NPHM  & 1.137 \\ 
opt. only w/ MICA     & \checkmark  & \checkmark  & $\mathscr{E}_{n}+\mathscr{E}_{\texttt{p}}$     & \checkmark       & NPHM  & 1.029 \\ \hline
Ours          & \checkmark  & \checkmark  & $\mathscr{E}_{n}+\mathscr{E}_{\texttt{p}}$     & \checkmark       & NPHM  & \textbf{0.988} \\ \bottomrule
\end{tabular}
}
\caption{\textbf{NoW Benchmark}~\cite{Sanyal2019now}: Ablations are performed on the validation subset of the NoW benchmark.}
\label{tab:abl_now}
\end{table}

\begin{table}[]
\centering
\setlength{\tabcolsep}{1.8pt}
\small
\begin{tabular}{@{}lrrrrrrrrr}
\toprule
     \multicolumn{1}{c}{\multirow{2}{*}{Method}}    & \multicolumn{3}{c}{Neutral}              && \multicolumn{3}{c}{Posed}                 \\
          \cmidrule(l){2-4}  \cmidrule(l){6-8}
         & L1\scriptsize{$\downarrow$}       & L2\scriptsize{$\downarrow$}        & NC\scriptsize{$\uparrow$}       &&  L1\scriptsize{$\downarrow$}       & L2\scriptsize{$\downarrow$}        & NC\scriptsize{$\uparrow$}     \\ \midrule

3D only & 1.64 & 1.11 &  0.894   && 1.65   & 1.11   & 0.890      \\

2D only & 1.79 & 1.21 &  0.893   && 1.67   & 1.13   & 0.893      \\
\hline
RGB input & 1.89 &  1.28 &  0.891  && 2.10 &  1.42 &  0.883    \\
Normals inp. & 1.76 &  1.19 &  0.893   && 1.87 &  1.26 &  0.886      \\
no $\mathscr{E}_{\texttt{p}}$ & 1.67 & 1.13 &  0.895   && 1.65 &  1.11 &  0.891      \\
\hline
ffwd. NPHM & 1.57 &  1.06 &  0.896    && 1.55 &  1.05 &  0.894 \\   
ffwd. FLAME & 1.71 & 1.15 &  0.883   && 1.81   & 1.22   & 0.881     \\
\hline
opt. only w/o MICA & 1.76   & 1.18   & 0.894 && 1.61 & 1.09 &  0.893        \\
opt. only w/ MICA & 1.60   & 1.08   & 0.890 && 1.50 & 1.01 &  0.895        \\
\hline
Ours     
    & \textbf{1.54} &  \textbf{1.04} &  \textbf{0.897}   && \textbf{1.37} &  \textbf{0.92} &\textbf{0.897} \\ \bottomrule
\end{tabular}
\caption{\textbf{Ablations on NeRSemble-SVFR Benchmark~\cite{giebenhain2025pixel3dmm}.}
}
\label{tab:abl_results}
\vspace{-0.3cm}
\end{table}
\newcolumntype{Y}{>{\centering\arraybackslash}X}
\newcolumntype{P}[1]{>{\centering\arraybackslash}p{#1}}
\begin{figure}[htb!]
    \centering
    \includegraphics[width=0.99\linewidth]{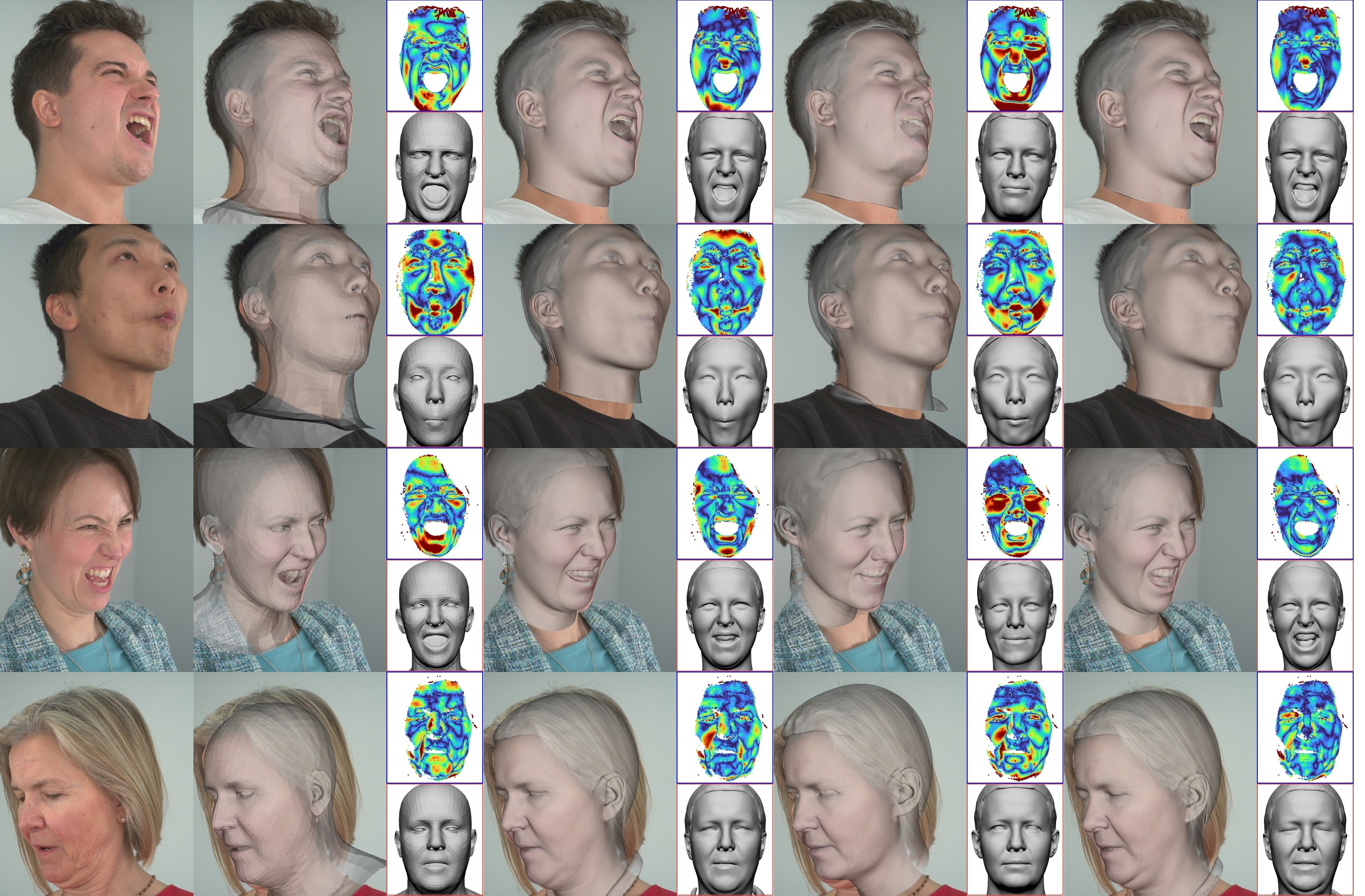}
    \setlength{\tabcolsep}{0pt}
    
    \begin{tabularx}{\linewidth}{P{0.18\linewidth}P{0.18\linewidth}YP{0.24\linewidth}Y}
        \scriptsize{Input} & 
        \scriptsize{ffwd. FLAME} & 
        \scriptsize{ffwd. NPHM} & 
        \scriptsize{Opt. only w/ MICA} & 
        \scriptsize{Ours}
    \end{tabularx}
    \caption{
    \textbf{Ablations, Posed:} NPHM feed-forward predictions exhibt more details compared to FLAME. Wihtout the feed-forward initialization our optimization sometimes fails to reconstruct extreme expressions (\eg see rows 1 and 3). 
    }
    \label{fig:alb_results_posed}
\end{figure}
\newcolumntype{Y}{>{\centering\arraybackslash}X}
\newcolumntype{P}[1]{>{\centering\arraybackslash}p{#1}}
\begin{figure}[htb!]
    \centering
    \includegraphics[width=0.99\linewidth]{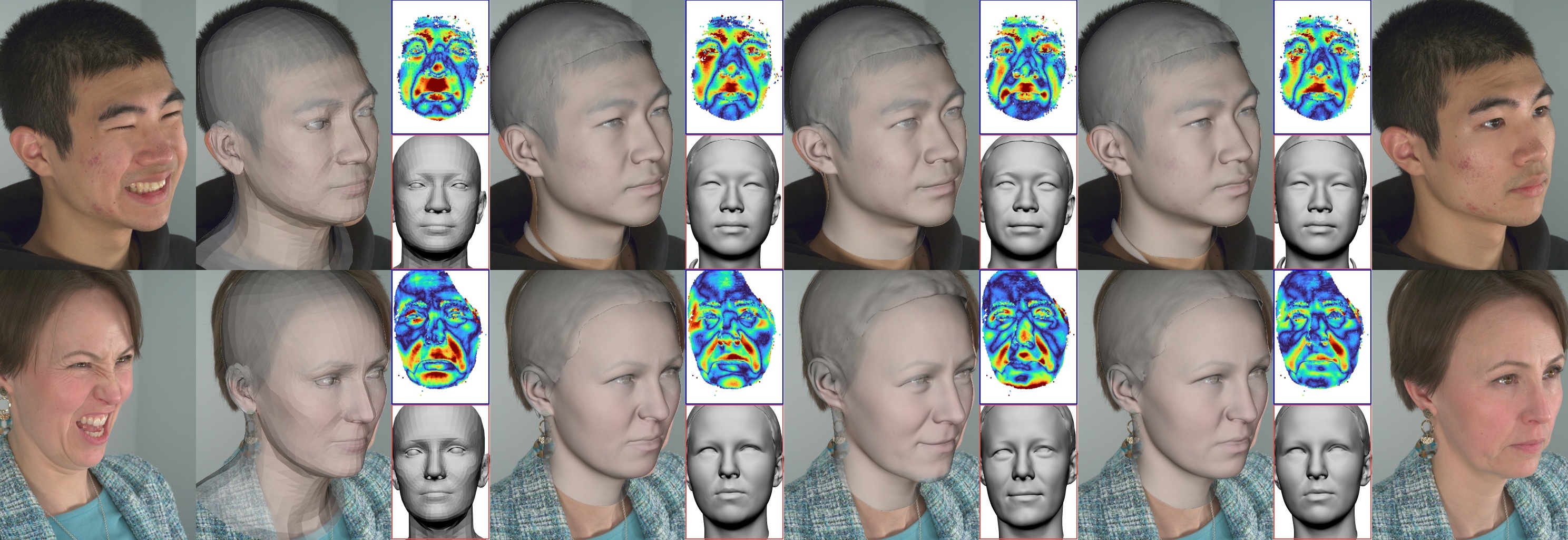}
    \setlength{\tabcolsep}{0pt}
    
\begin{tabularx}{\linewidth}{P{0.12\linewidth}P{0.18\linewidth}YP{0.24\linewidth}P{0.14\linewidth}P{0.12\linewidth}}
        \scriptsize{Input} & 
        \scriptsize{ffwd. FLAME~} & 
        \scriptsize~{~ffwd. NPHM} & 
        \scriptsize{Opt. only w/ MICA} & 
        \scriptsize{Ours} & 
                \scriptsize{Reference}
    \end{tabularx}
    \caption{
    \textbf{Ablations, Neutral:} Without the feed-forward prior, our optimization cannot properly disentangle identity and expression.
    }
    \label{fig:alb_results_neutral}
\end{figure}

 \subsection{Ablation Studies}
 \label{sec:ablations}

We demonstrate the effectiveness of our technical contributions by removing them one-by-one from our complete method. Quantitative and qualitative results are presented in \cref{fig:alb_results_posed,fig:alb_results_neutral}%
, and \cref{tab:abl_results,tab:abl_now}, respectively. Additionally, \cref{tab:abl_now} has a verbose description of the ablated components.

 \paragraph{What effect does the underlying 3DMM have?}
The main motivation for \OURS is the increased representational capacity of neural 3DMMs, \ie MonoNPHM in our case. To ablate the effect of the underlying 3DMM, we train a version of our approach with FLAME instead of MonoNPHM. 
Quantitative and qualitative results on the SVFR-Benchmark and NoW support our claim, that leveraging a neural 3DMM increases reconstruction fidelity.
Note that our FLAME-based feed-forward predictor outperforms the best available FLAME-based predictor MICA on the NoW validation set ($1.073$ vs. $1.109$ mean error), and outperforms SHeaP, the best FLAME-based feed-fordward method on the posed NeRSemble SVFR task.
On AffectNet, MonoNPHM also provides significantly better scores than FLAME, as shown in \cref{tab:affectnet}.

\paragraph{Inference-Time Optimization}
While our feed-forward prediction already reaches SotA performance, reconstruction fidelity can be improved further, \eg see \cref{fig:motivation}. Especially posed reconstructions benefit from optimization, while neutral reconstructions improve slightly (see \cref{tab:main_results}).
Importantly, we note that optimization without using our feed-forward prior sometimes fails to reconstruct complicated expressions, \eg see rows one and three of \cref{fig:alb_results_posed}.

\paragraph{Network Architecture}

Furthermore, we ablate several aspects of our method's architecture, especially \wrt to how the input image is encoded. 
Using DINOV2~\cite{oquab2023dinov2} encodings of the RGB input $I$, similar to Pixel3DMM~\cite{giebenhain2025pixel3dmm}, performs the worst, and did not properly converged on 2D training data. Using DINOv2 encodings of estimated normals $I^{\texttt{n}}{=}\mathscr{D}_{\texttt{n}}\left( \mathscr{E}_{\texttt{n}}(I)\right)$ resulted in similar performance on NoW, but increased performance on NeRSemble.
Directly using the tokens from $\mathscr{E}_{\texttt{n}}$ gives significantly better performance  and adding tokens from  $\mathscr{E}_{\texttt{p}}$ further boosts results.

\paragraph{Training Strategy/Data}
Finally, we show the importance of training on a mixture of 2D and 3D data. We hypothesis that 2D training data is essential to handle the appearance diversity of input images, while 3D data provides an unambiguous training signal.

\subsection{In-the-Wild Results}

For results on in-the-wild images we refer to \cref{fig:teaser} and our supplementary material.
Additionally, \cref{fig:motivation} compares our FLAME-based and NPHM-based regressors, and our full approach.

\section{Limitations and Future Work}
We believe that our scalable approach to 3D face reconstructions shows great potential, however, performance is currently limited by several aspects of MonoNPHM~\cite{giebenhain2024mononphm}. 
For example, its latent space prevents reliable registration of 3D hair styles, which leads to suboptimal geometry, even in the facial region. Moreover, the MLP-based volumetric rendering and marching cubes extraction introduce substantial computational overhead.
Future work could develop an improved NPHM variant, \eg leveraging 2DGS~\cite{Huang2DGS2024} for faster rendering and mesh extraction, and training on more diverse 3D hairstyles such as Difflocks~\cite{difflocks2025}.
Jointly fine-tuning NPHM with the feed-forward regressor is another promising direction.
Finally, as single-image reconstruction remains inherently ambiguous, incorporating uncertainty estimation, probabilistic reconstruction, \eg via conditional generation, or multi-image extensions, \eg by leveraging sequential information as in VGGT~\cite{wang2025vggt}, could resolve any remaining ambiguities.

\section{Conclusion}

We introduced \OURS, the first feed-forward framework for predicting NPHM parameters, enabling fast and high-fidelity 3D face reconstruction from a single input image.
Despite considerable research effort on FLAME parameter regression, our initial attempt on NPHM reconstruction demonstrates significant benefits.
Key to this success are our large-scale 3D data registration and the self-supervised training strategy on 2D data using surface normal estimators.
Moreover, we show that ViTs pretrained on face-specific geometric tasks capture facial structure far more effectively than DINOv2.
Together, these insights establish a new path toward scalable, high-quality monocular 3D face reconstruction.

\subsection*{Acknowledgements}
This work was supported by Toyota Motor Europe and Woven by Toyota.
This work was also supported by the ERC Consolidator Grant Gen3D (101171131).
We would also like to thank Angela Dai for the video voice-over.

{
    \small
    \bibliographystyle{ieeenat_fullname}
    \bibliography{main}

@String(CVPR= {IEEE Conf. Comput. Vis. Pattern Recog.})

@String(ICCV= {Int. Conf. Comput. Vis.})

@String(ECCV= {Eur. Conf. Comput. Vis.})

@String(TOG= {ACM Trans. Graph.})

@String(CVPR  = {CVPR})

@String(ICCV  = {ICCV})

@String(ECCV  = {ECCV})

@String(TOG   = {ACM TOG})

@inproceedings{blanz1999morphable,
  title={A morphable model for the synthesis of 3D faces},
  author={Blanz, Volker and Vetter, Thomas},
  booktitle={Proceedings of the 26th annual conference on Computer graphics and interactive techniques},
  pages={187--194},
  year={1999}
}

@inproceedings{giebenhain2024mononphm,
 author={Simon Giebenhain and Tobias Kirschstein and Markos Georgopoulos and  Martin R{\"{u}}nz and Lourdes Agapito and Matthias Nie{\ss}ner},
 title={MonoNPHM: Dynamic Head Reconstruction from Monocular Videos},
 booktitle = {Proc. IEEE Conf. on Computer Vision and Pattern Recognition (CVPR)},
 year = {2024}
}

@article{FLAME, 
  title = {Learning a model of facial shape and expression from {4D} scans}, 
  author = {Li, Tianye and Bolkart, Timo and Black, Michael. J. and Li, Hao and Romero, Javier}, 
  journal = {ACM Transactions on Graphics, (Proc. SIGGRAPH Asia)}, 
  volume = {36}, 
  number = {6}, 
  year = {2017}, 
  pages = {194:1--194:17},
  url = {https://doi.org/10.1145/3130800.3130813} 
}

@inproceedings{danvevcek2022emoca,
  title={Emoca: Emotion driven monocular face capture and animation},
  author={Dan{\v{e}}{\v{c}}ek, Radek and Black, Michael J and Bolkart, Timo},
  booktitle={Proceedings of the IEEE/CVF Conference on Computer Vision and Pattern Recognition},
  pages={20311--20322},
  year={2022}
}

@inproceedings{grassal2022neural,
    title={Neural head avatars from monocular RGB videos},
    author={Grassal, Philip-William and Prinzler, Malte and Leistner, Titus and 
            Rother, Carsten and Nie{\ss}ner, Matthias and Thies, Justus},
    booktitle={Proceedings of the IEEE/CVF Conference on Computer Vision and Pattern Recognition},
    pages={18653--18664},
    year={2022}
}

@inproceedings{Sanyal2019now,
title = {Learning to Regress {3D} Face Shape and Expression from an Image without {3D} Supervision},
author = {Sanyal, Soubhik and Bolkart, Timo and Feng, Haiwen and Black, Michael},
booktitle = {Proceedings IEEE Conf. on Computer Vision and Pattern Recognition (CVPR)},
month = jun,
pages = {7763--7772},
year = {2019},
month_numeric = {6} 
}

@article{zhu2023facescape,
  title={FaceScape: 3D Facial Dataset and Benchmark for Single-View 3D Face Reconstruction},
  author={Zhu, Hao and Yang, Haotian and Guo, Longwei and Zhang, Yidi and Wang, Yanru and Huang, Mingkai and Wu, Menghua 
  and Shen, Qiu and Yang, Ruigang and Cao, Xun},
   journal={IEEE Transactions on Pattern Analysis and Machine Intelligence (TPAMI)},
  year={2023},
  publisher={IEEE}}

@inproceedings{zielonka2022towards, 
  title={Towards metrical reconstruction of human faces},
  author={Zielonka, Wojciech and Bolkart, Timo and Thies, Justus},
  booktitle={European conference on computer vision},
  pages={250--269},
  year={2022},
  organization={Springer}
}

@InProceedings{taubner2024flowface,
  author    = {Taubner, Felix and Raina, Prashant and Tuli, Mathieu and Teh, Eu Wern and Lee, Chul and Huang, Jinmiao},
  title     = {{3D} Face Tracking from {2D} Video through Iterative Dense {UV} to Image Flow},
  booktitle = {Proceedings of the IEEE/CVF Conference on Computer Vision and Pattern Recognition (CVPR)},
  month     = {June},
  year      = {2024},
  pages     = {1227-1237}
}

@article{feng2021learning_deca,
  title={Learning an animatable detailed 3D face model from in-the-wild images},
  author={Feng, Yao and Feng, Haiwen and Black, Michael J and Bolkart, Timo},
  journal={ACM Transactions on Graphics (ToG)},
  volume={40},
  number={4},
  pages={1--13},
  year={2021},
  publisher={ACM New York, NY, USA}
}

@misc{filntisis2022visual,
  title = {Visual Speech-Aware Perceptual 3D Facial Expression Reconstruction from Videos},
  author = {Filntisis, Panagiotis P. and Retsinas, George and Paraperas-Papantoniou, Foivos and Katsamanis, Athanasios and Roussos, Anastasios and Maragos, Petros},
  publisher = {arXiv},
  year = {2022},
}

@inproceedings{SMIRK:CVPR:2024,
    title = {3D Facial Expressions through Analysis-by-Neural-Synthesis},
    author = {Retsinas, George and Filntisis, Panagiotis P. and Danecek, Radek and Abrevaya, Victoria F. and Roussos, Anastasios and Bolkart, Timo and Maragos, Petros},
    booktitle = {Conference on Computer Vision and Pattern Recognition (CVPR)},
    year = {2024}
}

@inproceedings{zhang2023accurate,
  title={Accurate 3d face reconstruction with facial component tokens},
  author={Zhang, Tianke and Chu, Xuangeng and Liu, Yunfei and Lin, Lijian and Yang, Zhendong and Xu, Zhengzhuo and Cao, Chengkun and Yu, Fei and Zhou, Changyin and Yuan, Chun and others},
  booktitle={Proceedings of the IEEE/CVF international conference on computer vision},
  pages={9033--9042},
  year={2023}
}

@inproceedings{wood2022denselandmarks,
  title={3d face reconstruction with dense landmarks},
  author={Wood, Erroll and Baltru{\v{s}}aitis, Tadas and Hewitt, Charlie and Johnson, Matthew and Shen, Jingjing and Milosavljevi{\'c}, Nikola and Wilde, Daniel and Garbin, Stephan and Sharp, Toby and Stojiljkovi{\'c}, Ivan and others},
  booktitle={European Conference on Computer Vision},
  pages={160--177},
  year={2022},
  organization={Springer}
}

@inproceedings{guo20203ddfav2,
  title={Towards fast, accurate and stable 3d dense face alignment},
  author={Guo, Jianzhu and Zhu, Xiangyu and Yang, Yang and Yang, Fan and Lei, Zhen and Li, Stan Z},
  booktitle={European Conference on Computer Vision},
  pages={152--168},
  year={2020},
  organization={Springer}
}

@inproceedings{thies2016face2face,
  title={Face2face: Real-time face capture and reenactment of rgb videos},
  author={Thies, Justus and Zollhofer, Michael and Stamminger, Marc and Theobalt, Christian and Nie{\ss}ner, Matthias},
  booktitle={Proceedings of the IEEE conference on computer vision and pattern recognition},
  pages={2387--2395},
  year={2016}
}

@inproceedings{giebenhain2023nphm,
  title={Learning neural parametric head models},
  author={Giebenhain, Simon and Kirschstein, Tobias and Georgopoulos, Markos and R{\"u}nz, Martin and Agapito, Lourdes and Nie{\ss}ner, Matthias},
  booktitle={Proceedings of the IEEE/CVF Conference on Computer Vision and Pattern Recognition},
  pages={21003--21012},
  year={2023}
}

@article{oquab2023dinov2,
  title={Dinov2: Learning robust visual features without supervision},
  author={Oquab, Maxime and Darcet, Timoth{\'e}e and Moutakanni, Th{\'e}o and Vo, Huy and Szafraniec, Marc and Khalidov, Vasil and Fernandez, Pierre and Haziza, Daniel and Massa, Francisco and El-Nouby, Alaaeldin and others},
  journal={arXiv preprint arXiv:2304.07193},
  year={2023}
}

@article{dosovitskiy2020vit,
  title={An image is worth 16x16 words: Transformers for image recognition at scale},
  author={Dosovitskiy, Alexey and Beyer, Lucas and Kolesnikov, Alexander and Weissenborn, Dirk and Zhai, Xiaohua and Unterthiner, Thomas and Dehghani, Mostafa and Minderer, Matthias and Heigold, Georg and Gelly, Sylvain and others},
  journal={arXiv preprint arXiv:2010.11929},
  year={2020}
}

@misc{adam,
  author = {Kingma, Diederik P. and Ba, Jimmy},
  description = {[1412.6980] Adam: A Method for Stochastic Optimization},
  note = {cite arxiv:1412.6980Comment: Published as a conference paper at the 3rd International Conference  for Learning Representations, San Diego, 2015},
  title = {Adam: A Method for Stochastic Optimization},
  url = {http://arxiv.org/abs/1412.6980},
  year = 2014
}

@inproceedings{wang2024dust3r,
  title={Dust3r: Geometric 3d vision made easy},
  author={Wang, Shuzhe and Leroy, Vincent and Cabon, Yohann and Chidlovskii, Boris and Revaud, Jerome},
  booktitle={Proceedings of the IEEE/CVF Conference on Computer Vision and Pattern Recognition},
  pages={20697--20709},
  year={2024}
}

@inproceedings{kirschstein2024diffusionavatars,
  title={Diffusionavatars: Deferred diffusion for high-fidelity 3d head avatars},
  author={Kirschstein, Tobias and Giebenhain, Simon and Nie{\ss}ner, Matthias},
  booktitle={Proceedings of the IEEE/CVF Conference on Computer Vision and Pattern Recognition},
  pages={5481--5492},
  year={2024}
}

@article{prinzler2024joker,
  title={Joker: Conditional 3D Head Synthesis with Extreme Facial Expressions},
  author={Prinzler, Malte and Zakharov, Egor and Sklyarova, Vanessa and Kabadayi, Berna and Thies, Justus},
  journal={arXiv preprint arXiv:2410.16395},
  year={2024}
}

@article{taubner2024cap4d,
  title={CAP4D: Creating Animatable 4D Portrait Avatars with Morphable Multi-View Diffusion Models},
  author={Taubner, Felix and Zhang, Ruihang and Tuli, Mathieu and Lindell, David B},
  journal={arXiv preprint arXiv:2412.12093},
  year={2024}
}

@misc{giebenhain2025pixel3dmm,
title={Pixel3DMM: Versatile Screen-Space Priors for Single-Image 3D Face Reconstruction},
author={Simon Giebenhain and Tobias Kirschstein and  Martin R{\"{u}}nz and Lourdes Agapito and Matthias Nie{\ss}ner},
year={2025},
url={https://arxiv.org/abs/2505.00615},
}

@InProceedings{Schoneveld_2025_ICCV,
    author    = {Schoneveld, Liam and Chen, Zhe and Davoli, Davide and Tang, Jiapeng and Terazawa, Saimon and Nishino, Ko and Nie{\ss}ner, Matthias},
    title     = {SHeaP: Self-Supervised Head Geometry Predictor Learned via 2D Gaussians},
    booktitle = {Proceedings of the IEEE/CVF International Conference on Computer Vision (ICCV)},
    month     = {October},
    year      = {2025},
    pages     = {14162-14172}
}

@article{wang2021neus,
      title={NeuS: Learning Neural Implicit Surfaces by Volume Rendering for Multi-view Reconstruction}, 
      author={Peng Wang and Lingjie Liu and Yuan Liu and Christian Theobalt and Taku Komura and Wenping Wang},
	  journal={NeurIPS},
      year={2021}
}

@misc{saleh2025david,
    title={{DAViD}: Data-efficient and Accurate Vision Models from Synthetic Data},
    author={Fatemeh Saleh and Sadegh Aliakbarian and Charlie Hewitt and Lohit Petikam and Xiao-Xian and Antonio Criminisi and Thomas J. Cashman and Tadas Baltrušaitis},
    year={2025},
    eprint={2507.15365},
    archivePrefix={arXiv},
    primaryClass={cs.CV},
    url={https://arxiv.org/abs/2507.15365},
}

@inproceedings{thanos2022_mimicme,
  title={MimicME: A Large Scale Diverse 4D Database for Facial Expression Analysis},
  author={Papaioannou, Athanasios an Baris, Gecer and Cheng, Shiyang and   Chrysos, Grigorios G. and Deng, Jiankang and Fotiadou, Eftychia and  Kampouris, Christos and Kollias, Dimitrios and  Moschoglou, Stylianos and  Songsri-In, Kritaphat and Ploumpis, Stylianos and Trigeorgis, George and Tzirakis, Panagiotis and Ververas, Evangelos and Zhou, Yuxiang and  Ponniah, Allan and Roussos, Anastasios and Zafeiriou, Stefanos},  
booktitle={Proceedings of the European Conference on Computer Vision},
  year={2022}
}

@article{dai2020statistical_lyhm,
  title={Statistical Modeling of Craniofacial Shape and Texture},
  author={Dai, Hang and Pears, Nick and Smith, William A. P. and Duncan, Christian},
  journal={International Journal of Computer Vision},
  volume={128},
  number={2},
  pages={547--571},
  year={2020}
}

@misc{cui2024hallo3,
	title={Hallo3: Highly Dynamic and Realistic Portrait Image Animation with Video Diffusion Transformer}, 
	author={Jiahao Cui and Hui Li and Yun Zhan and Hanlin Shang and Kaihui Cheng and Yuqi Ma and Shan Mu and Hang Zhou and Jingdong Wang and Siyu Zhu},
	year={2024},
	eprint={2412.00733},
	archivePrefix={arXiv},
	primaryClass={cs.CV}
}

@inproceedings{zhu2022celebvhq,
  title={{CelebV-HQ}: A Large-Scale Video Facial Attributes Dataset},
  author={Zhu, Hao and Wu, Wayne and Zhu, Wentao and Jiang, Liming and Tang, Siwei and Zhang, Li and Liu, Ziwei and Loy, Chen Change},
  booktitle={ECCV},
  year={2022}
}

@inproceedings{yu2022celebvtext,
  title={{CelebV-Text}: A Large-Scale Facial Text-Video Dataset},
  author={Yu, Jianhui and Zhu, Hao and Jiang, Liming and Loy, Chen Change and Cai, Weidong and Wu, Wayne},
  booktitle={CVPR},
  year={2023}
}

@inproceedings{qian2024gaussianavatars,
  title={Gaussianavatars: Photorealistic head avatars with rigged 3d gaussians},
  author={Qian, Shenhan and Kirschstein, Tobias and Schoneveld, Liam and Davoli, Davide and Giebenhain, Simon and Nie{\ss}ner, Matthias},
  booktitle={Proceedings of the IEEE/CVF Conference on Computer Vision and Pattern Recognition},
  pages={20299--20309},
  year={2024}
}

@inproceedings{xu2024gphm,
                title={3D Gaussian Parametric Head Model},
                author={Xu, Yuelang and Wang, Lizhen and Zheng, Zerong and Su, Zhaoqi and Liu, Yebin},
                booktitle={Proceedings of the European Conference on Computer Vision (ECCV)},
                year={2024}
              }

@article{palafox2021npms,
    author        = {Palafox, Pablo and Bo{\v{z}}i{\v{c}}, Alja{\v{z}} and Thies, Justus and Nie{\ss}ner, Matthias and Dai, Angela},
    title         = {NPMs: Neural Parametric Models for 3D Deformable Shapes},
    journal       = {arXiv preprint arXiv:2104.00702},
    year          = {2021},
}

@inproceedings{zheng2022imface,
  title={ImFace: A Nonlinear 3D Morphable Face Model with Implicit Neural Representations},
  author={Zheng, Mingwu and Yang, Hongyu and Huang, Di and Chen, Liming},
  booktitle={Proceedings of the IEEE/CVF Conference on Computer Vision and Pattern Recognition},
  pages={20343--20352},
  year={2022}
}

@inproceedings{yenamandra2021i3dmm,
  title={i3DMM: Deep Implicit 3D Morphable Model of Human Heads},
  author={Yenamandra, Tarun and Tewari, Ayush and Bernard, Florian and Seidel, Hans-Peter and Elgharib, Mohamed and Cremers, Daniel and Theobalt, Christian},
  booktitle={Proceedings of the IEEE/CVF Conference on Computer Vision and Pattern Recognition},
  pages={12803--12813},
  year={2021}
}

@PROCEEDINGS{bfm09,
            title={A 3D Face Model for Pose and Illumination Invariant Face Recognition},
            author={P. Paysan and R. Knothe and B. Amberg
                    and S. Romdhani and T. Vetter},
            journal={Proceedings of the 6th IEEE International Conference on Advanced Video and Signal based Surveillance (AVSS)
                 for Security, Safety and Monitoring in Smart Environments},
            organization={IEEE},
            year={2009},
            address     = {Genova, Italy},
            }

@article{mollahosseini2017affectnet,
  title={Affectnet: A database for facial expression, valence, and arousal computing in the wild},
  author={Mollahosseini, Ali and Hasani, Behzad and Mahoor, Mohammad H},
  journal={IEEE Transactions on Affective Computing},
  volume={10},
  number={1},
  pages={18--31},
  year={2017},
  publisher={IEEE}
}

@inproceedings{richardson2017learning,
  title={Learning detailed face reconstruction from a single image},
  author={Richardson, Elad and Sela, Matan and Or-El, Roy and Kimmel, Ron},
  booktitle={Proceedings of the IEEE conference on computer vision and pattern recognition},
  pages={1259--1268},
  year={2017}
}

@inproceedings{richardson20163d,
  title={3D face reconstruction by learning from synthetic data},
  author={Richardson, Elad and Sela, Matan and Kimmel, Ron},
  booktitle={2016 fourth international conference on 3D vision (3DV)},
  pages={460--469},
  year={2016},
  organization={IEEE}
}

@inproceedings{tuan2017regressing,
  title={Regressing robust and discriminative 3D morphable models with a very deep neural network},
  author={Tuan Tran, Anh and Hassner, Tal and Masi, Iacopo and Medioni, G{\'e}rard},
  booktitle={Proceedings of the IEEE conference on computer vision and pattern recognition},
  pages={5163--5172},
  year={2017}
}

@inproceedings{trần2018extreme,
  title={Extreme 3d face reconstruction: Seeing through occlusions},
  author={Tran, Anh Tuan and Hassner, Tal and Masi, Iacopo and Paz, Eran and Nirkin, Yuval and Medioni, G{\'e}rard},
  booktitle={Proceedings of the IEEE conference on computer vision and pattern recognition},
  pages={3935--3944},
  year={2018}
}

@inproceedings{tewari2017mofa,
  title={Mofa: Model-based deep convolutional face autoencoder for unsupervised monocular reconstruction},
  author={Tewari, Ayush and Zollhofer, Michael and Kim, Hyeongwoo and Garrido, Pablo and Bernard, Florian and Perez, Patrick and Theobalt, Christian},
  booktitle={Proceedings of the IEEE international conference on computer vision workshops},
  pages={1274--1283},
  year={2017}
}

@inproceedings{wang2025vggt,
  title={VGGT: Visual Geometry Grounded Transformer},
  author={Wang, Jianyuan and Chen, Minghao and Karaev, Nikita and Vedaldi, Andrea and Rupprecht, Christian and Novotny, David},
  booktitle={Proceedings of the IEEE/CVF Conference on Computer Vision and Pattern Recognition},
  year={2025}
}

@inproceedings{difflocks2025,
  title = {{DiffLocks}: Generating 3D Hair from a Single Image using Diffusion Models},
  author = {Rosu, Radu Alexandru and Wu, Keyu and Feng, Yao and Zheng, Youyi and Black, Michael J.},
  booktitle = {IEEE/CVF Conf.~on Computer Vision and Pattern Recognition(CVPR)},
  year = {2025}
}

@inproceedings{Huang2DGS2024,
    title={2D Gaussian Splatting for Geometrically Accurate Radiance Fields},
    author={Huang, Binbin and Yu, Zehao and Chen, Anpei and Geiger, Andreas and Gao, Shenghua},
    publisher = {Association for Computing Machinery},
    booktitle = {SIGGRAPH 2024 Conference Papers},
    year      = {2024},
    doi       = {10.1145/3641519.3657428}
}

@inproceedings{aneja2023facetalk,
      author={Shivangi Aneja and Justus Thies and Angela Dai and Matthias Nießner},
      title={FaceTalk: Audio-Driven Motion Diffusion for Neural Parametric Head Models}, 
      booktitle = {Proc. IEEE Conf. on Computer Vision and Pattern Recognition (CVPR)},
      year={2024},
}

@inproceedings{giebenhain2024npga,
 author    = {Simon Giebenhain and Tobias Kirschstein and  Martin R{\"{u}}nz and Lourdes Agapito and Matthias Nie{\ss}ner},
 title     = {NPGA: Neural Parametric Gaussian Avatars},
 booktitle = {SIGGRAPH Asia 2024 Conference Papers (SA Conference Papers '24), December 3-6, Tokyo, Japan},
 doi       = {10.1145/3680528.3687689},
 isbn      = {979-8-4007-1131-2/24/12},
 year      = {2024},
}

@Article{kerbl3Dgaussians,
      author       = {Kerbl, Bernhard and Kopanas, Georgios and Leimk{\"u}hler, Thomas and Drettakis, George},
      title        = {3D Gaussian Splatting for Real-Time Radiance Field Rendering},
      journal      = {ACM Transactions on Graphics},
      number       = {4},
      volume       = {42},
      month        = {July},
      year         = {2023},
      url          = {https://repo-sam.inria.fr/fungraph/3d-gaussian-splatting/}
}

@inproceedings{li2024uravatar,
          author = {Junxuan Li and Chen Cao and Gabriel Schwartz and Rawal Khirodkar and Christian Richardt and Tomas Simon and Yaser Sheikh and Shunsuke Saito},
          title = {URAvatar: Universal Relightable Gaussian Codec Avatars}, 
          booktitle = {ACM SIGGRAPH 2024 Conference Papers},
          year = {2024},
        }

@InProceedings{Kirschstein_2025_ICCV_avat3r,
    author    = {Kirschstein, Tobias and Romero, Javier and Sevastopolsky, Artem and Nie{\ss}ner, Matthias and Saito, Shunsuke},
    title     = {Avat3r: Large Animatable Gaussian Reconstruction Model for High-fidelity 3D Head Avatars},
    booktitle = {Proceedings of the IEEE/CVF International Conference on Computer Vision (ICCV)},
    month     = {October},
    year      = {2025},
    pages     = {12089-12100}
}

@InProceedings{ding2023diffusionrig,
      author    = {Zheng Ding,and Cecilia Zhang and Zhihao Xia and Lars Jebe and Zhuowen Tu and Xiuming Zhang},
      title     = {DiffusionRig: Learning Personalized Priors for Facial Appearance Editing},
      booktitle = {Proceedings of the IEEE/CVF Conference on Computer Vision and Pattern Recognition},
      year      = {2023},
}

@misc{taubner2025mvp4d,
  title={{MVP4D}: Multi-View Portrait Video Diffusion for Animatable {4D} Avatars}, 
  author={Felix Taubner and Ruihang Zhang and Mathieu Tuli and Sherwin Bahmani and David B. Lindell},
  year={2025},
  eprint={2510.12785},
  archivePrefix={arXiv},
  primaryClass={cs.CV},
  url={https://arxiv.org/abs/2510.12785}, 
}

@article{zheng2023imface++,
  title={ImFace++: A sophisticated nonlinear 3D morphable face model with implicit neural representations},
  author={Zheng, Mingwu and Zhang, Haiyu and Yang, Hongyu and Chen, Liming and Huang, Di},
  journal={IEEE Transactions on Pattern Analysis and Machine Intelligence},
  year={2024},
  publisher={IEEE}
}

@inproceedings{chen2022gdna,
  title={gdna: Towards generative detailed neural avatars},
  author={Chen, Xu and Jiang, Tianjian and Song, Jie and Yang, Jinlong and Black, Michael J and Geiger, Andreas and Hilliges, Otmar},
  booktitle={Proceedings of the IEEE/CVF Conference on Computer Vision and Pattern Recognition},
  pages={20427--20437},
  year={2022}
}

@article{auth_vol_ava,
author = {Cao, Chen and Simon, Tomas and Kim, Jin Kyu and Schwartz, Gabe and Zollhoefer, Michael and Saito, Shunsuke and Lombardi, Stephen and Wei, Shih-En and Belko, Danielle and Yu, Shoou-I and Sheikh, Yaser and Saragih, Jason},
title = {Authentic volumetric avatars from a phone scan},
year = {2022},
issue_date = {July 2022},
publisher = {Association for Computing Machinery},
address = {New York, NY, USA},
volume = {41},
number = {4},
issn = {0730-0301},
url = {https://doi.org/10.1145/3528223.3530143},
doi = {10.1145/3528223.3530143},
journal = {ACM Trans. Graph.},
month = jul,
articleno = {163},
numpages = {19}
}

@inproceedings{morf,
author = {Wang, Daoye and Chandran, Prashanth and Zoss, Gaspard and Bradley, Derek and Gotardo, Paulo},
title = {MoRF: Morphable Radiance Fields for Multiview Neural Head Modeling},
year = {2022},
isbn = {9781450393379},
publisher = {Association for Computing Machinery},
address = {New York, NY, USA},
url = {https://doi.org/10.1145/3528233.3530753},
doi = {10.1145/3528233.3530753},
booktitle = {ACM SIGGRAPH 2022 Conference Proceedings},
articleno = {55},
numpages = {9},
location = {Vancouver, BC, Canada},
series = {SIGGRAPH '22}
}

@article{mildenhall2021nerf,
  title={Nerf: Representing scenes as neural radiance fields for view synthesis},
  author={Mildenhall, Ben and Srinivasan, Pratul P and Tancik, Matthew and Barron, Jonathan T and Ramamoorthi, Ravi and Ng, Ren},
  journal={Communications of the ACM},
  volume={65},
  number={1},
  pages={99--106},
  year={2021},
  publisher={ACM New York, NY, USA}
}

@inproceedings{marchingcubes,
author = {Lorensen, William E. and Cline, Harvey E.},
title = {Marching cubes: A high resolution 3D surface construction algorithm},
year = {1987},
isbn = {0897912276},
publisher = {Association for Computing Machinery},
address = {New York, NY, USA},
url = {https://doi.org/10.1145/37401.37422},
doi = {10.1145/37401.37422},
booktitle = {Proceedings of the 14th Annual Conference on Computer Graphics and Interactive Techniques},
pages = {163–169},
numpages = {7},
series = {SIGGRAPH '87}
}

@article{zheng2024headgap,
  title={HeadGAP: Few-shot 3D Head Avatar via Generalizable Gaussian Priors},
  author={Zheng, Xiaozheng and Wen, Chao and Li, Zhaohu and Zhang, Weiyi and Su, Zhuo and Chang, Xu and Zhao, Yang and Lv, Zheng and Zhang, Xiaoyuan and Zhang, Yongjie and Wang, Guidong and Xu Lan},
  journal={arXiv preprint arXiv:2408.06019},
  year={2024}
}

@incollection{buehler2024cafca,
    title={Cafca: High-quality Novel View Synthesis of Expressive Faces from Casual Few-shot Captures},
    author={Marcel C. Buehler and Gengyan Li and Erroll Wood and Leonhard Helminger and Xu Chen and Tanmay Shah and Daoye Wang and Stephan Garbin and Sergio Orts-Escolano and Otmar Hilliges and Dmitry Lagun and Jérémy Riviere and Paulo Gotardo and Thabo Beeler and Abhimitra Meka and Kripasindhu Sarkar},
    year={2024},
    booktitle={ACM SIGGRAPH Asia 2024 Conference Paper},
    doi={10.1145/3680528.3687580},
    url={https://doi.org/10.1145/3680528}
}
}

\end{document}